\documentclass[runningheads]{llncs}
\usepackage{hyperref} 
\hypersetup{
    colorlinks=true,
    linkcolor=blue,
    citecolor=blue
}
\usepackage[T1]{fontenc}
\usepackage{graphicx}
\usepackage{bm} 
\usepackage{amsmath}
\usepackage{graphicx}
\usepackage{multirow}
\usepackage{tabularx}
\usepackage{booktabs}       
\usepackage{pifont}
\usepackage{amsfonts}       
\usepackage{caption}
\usepackage{subcaption}
\usepackage{xcolor}         
\definecolor{green}{rgb}{0.0, 0.5, 0.0}
\definecolor{yellow}{rgb}{0.8, 0.8, 0}
\definecolor{cyan}{rgb}{0, 0.8, 0.8}
\usepackage{marvosym}

\begin{document}

\title{Unsupervised Domain Adaptation for Anatomical Landmark Detection}

\author{Haibo Jin$^{1}$, Haoxuan Che$^{1}$, \and Hao Chen\textsuperscript{\Letter}$^{1,2}$ \\
}

\institute{$^{1}$Department of Computer Science and Engineering\\
$^{2}$Department of Chemical and Biological Engineering\\
The Hong Kong University of Science and Technology, Kowloon, Hong Kong\\
\texttt{ \{hjinag,hche,jhc\}@cse.ust.hk}}

\maketitle              
\begin{abstract}
Recently, anatomical landmark detection has achieved great progresses on single-domain data, which usually assumes training and test sets are from the same domain. However, such an assumption is not always true in practice, which can cause significant performance drop due to domain shift. To tackle this problem, we propose a novel framework for anatomical landmark detection under the setting of unsupervised domain adaptation (UDA), which aims to transfer the knowledge from labeled source domain to unlabeled target domain. The framework leverages \textit{self-training} and \textit{domain adversarial learning} to address the domain gap during adaptation. Specifically, a self-training strategy is proposed to select reliable landmark-level pseudo-labels of target domain data with dynamic thresholds, which makes the adaptation more effective. Furthermore, a domain adversarial learning module is designed to handle the unaligned data distributions of two domains by learning domain-invariant features via adversarial training. Our experiments on cephalometric and lung landmark detection show the effectiveness of the method, which reduces the domain gap by a large margin and outperforms other UDA methods consistently. The code is available at \href{https://github.com/jhb86253817/UDA\_Med\_Landmark}{https://github.com/jhb86253817/UDA\_Med\_Landmark}.
\end{abstract}

\section{Introduction}
\label{sec1}

Anatomical landmark detection is a fundamental step in many clinical applications such as orthodontic diagnosis~\cite{JLW22} and orthognathic treatment planning~\cite{CMC19}. However, manually locating the landmarks can be tedious and time-consuming. And the results from manual labeling can cause errors due to the inconsistency in landmark identification~\cite{CCK14}. Therefore, it is of great need to automate the task of landmark detection for efficiency and consistency.    

In recent years, deep learning based methods have achieved great progresses in anatomical landmark detection. For supervised learning, earlier works~\cite{PSB19,ZLZ19,CMC19} adopted heatmap regression with extra shape constraints. Later, graph network~\cite{LLZ20} and self-attention~\cite{JLW22} were introduced to model landmark dependencies in an end-to-end manner for better performance. There are also other works explored one-shot~\cite{YQX21}, universal~\cite{ZYX21}, 3D~\cite{LCD22,ELF22}, and interactive models~\cite{KKK22}. 

\begin{figure}[t]
\centering
  \includegraphics[width=1\linewidth]{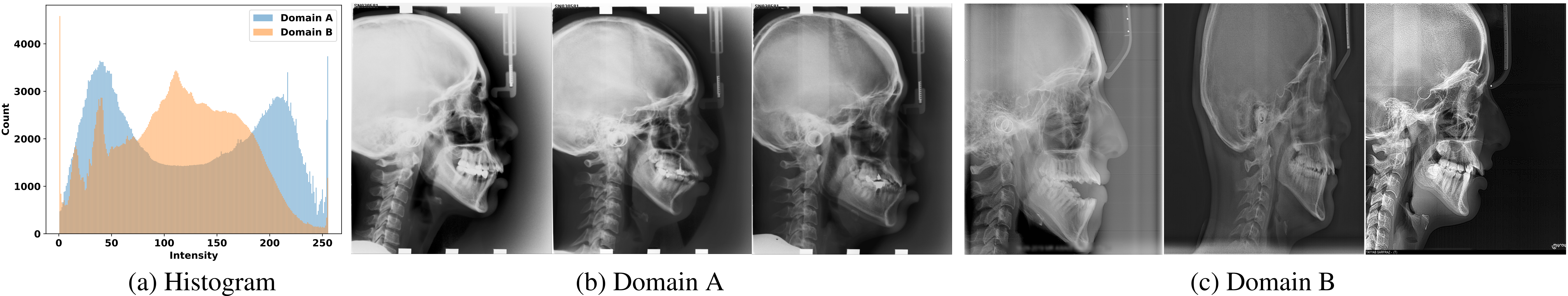}
\vspace{-2pt}
\caption{Domain A vs. Domain B. (a) Image histogram. (b)-(c) Visual samples.} 
\label{fig:domain_shift}      
\end{figure}

Despite the success of recent methods, they mostly focus on single-domain data, which assume the training and test sets follow the same distribution. However, such an assumption is not always true in practice, due to the differences in patient populations and imaging devices. Fig.~\ref{fig:domain_shift} shows that cephalogram images from two domains can be very different in both histogram and visual appearance. Therefore, a well trained model may encounter severe performance degradation in practice due to the domain shift of test data. A straightforward solution to this issue is to largely increase the size and diversity of training set, but the labeling is prohibitively expensive, especially for medical images. On the other hand, unsupervised domain adaptation (UDA)~\cite{GaL15} aims to transfer the knowledge learned from the labeled source domain to the unlabeled target domain, which is a potential solution to the domain shift problem~\cite{CJC22,CCJ23} as unlabeled data is much easier to collect. The effectiveness of UDA has been proven in many vision tasks, such as image classification~\cite{GaL15}, object detection~\cite{CLS18}, person re-identification~\cite{WLW19}, and pose estimation~\cite{YOW18,BHD22,MQH20}. However, its feasibility in anatomical landmark detection still remains unknown.
 
In this paper, we aim to investigate anatomical landmark detection under the setting of UDA. Our preliminary experiments show that a well-performed model will yield significant performance drop on cross-domain data, where the mean radial error (MRE) increases from 1.22mm to 3.32mm and the success detection rate (SDR) within 2mm drops from 83.76\% to 50.05\%. To address the domain gap, we propose a unified framework, which contains a base landmark detection model, a self-training strategy, and a domain adversarial learning module. Specifically, self-training is adopted to effectively leverage the unlabeled data from the target domain via pseudo-labels. To handle confirmation bias~\cite{AOA20}, we propose landmark-aware self-training (LAST) to select pseudo-labels at the landmark-level with dynamic thresholds. Furthermore, to address the covariate shift~\cite{ZYZ22} issue (i.e., unaligned data distribution) that may degrade the performance of self-training, a domain adversarial learning (DAL) module is designed to learn domain-invariant features via adversarial training. Our experiments on two anatomical datasets show the effectiveness of the proposed framework. For example, on cephalometric landmark detection, it reduces the domain gap in MRE by 47\% (3.32mm$\rightarrow$1.75mm) and improves the SDR (2mm) from 50.05\% to 69.15\%. We summarize our contributions as follows.
\begin{enumerate}
\item We investigated anatomical landmark detection under the UDA setting for the \textit{first time}, and showed that domain shift indeed causes severe performance drop of a well-performed landmark detection model.  
\item We proposed a novel framework for the UDA of anatomical landmark detection, which significantly improves the cross-domain performance and consistently outperforms other state-of-the-art UDA methods.  
\end{enumerate}   

\begin{figure}[t]
\centering
  \includegraphics[width=0.9\linewidth]{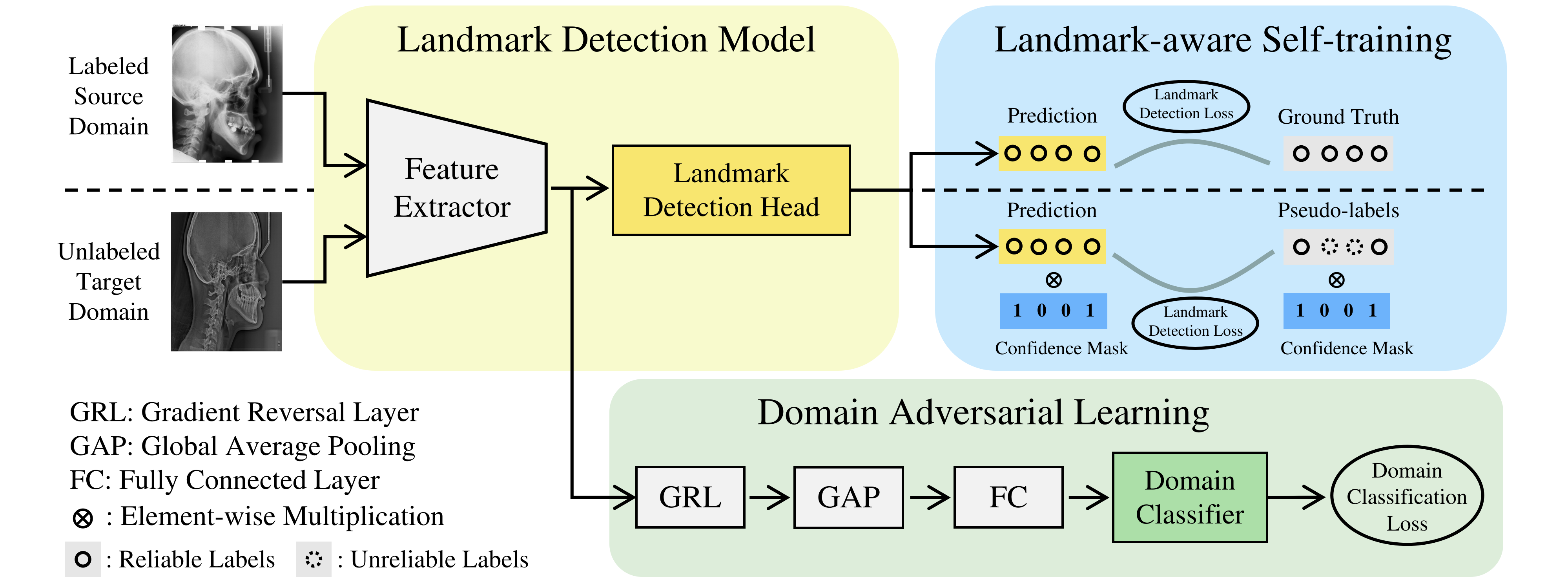}
\caption{The overall framework. Based on 1) the landmark detection model, it 2) utilizes LAST to leverage the unlabeled target domain data via pseudo-labels, and 3) simultaneously conducts DAL for domain-invariant features.} 
\label{fig:framework}      
\end{figure}

\section{Method}
\label{sec2}
Fig.~\ref{fig:framework} shows the overall framework, which aims to yield satisfactory performance in target domain under the UDA setting. During training, it leverages both labeled source domain data $\mathcal{S}=\{x^\mathcal{S}_i,y^\mathcal{S}_i\}^N_{i=1}$ and unlabeled target domain data $\mathcal{T}=\{x^\mathcal{T}_j\}^M_{j=1}$. For evaluation, it will be tested on a hold-out test set from target domain. The landmark detection model is able to predict landmarks with confidence, which is detailed in Sec.~\ref{sec2.1}. To reduce domain gap, we further propose LAST and DAL, which are introduced in Sec.~\ref{sec2.2} and \ref{sec2.3}, respectively.

\subsection{Landmark Detection Model}
\label{sec2.1}
\begin{figure}[t]
\centering
  \includegraphics[width=1\linewidth]{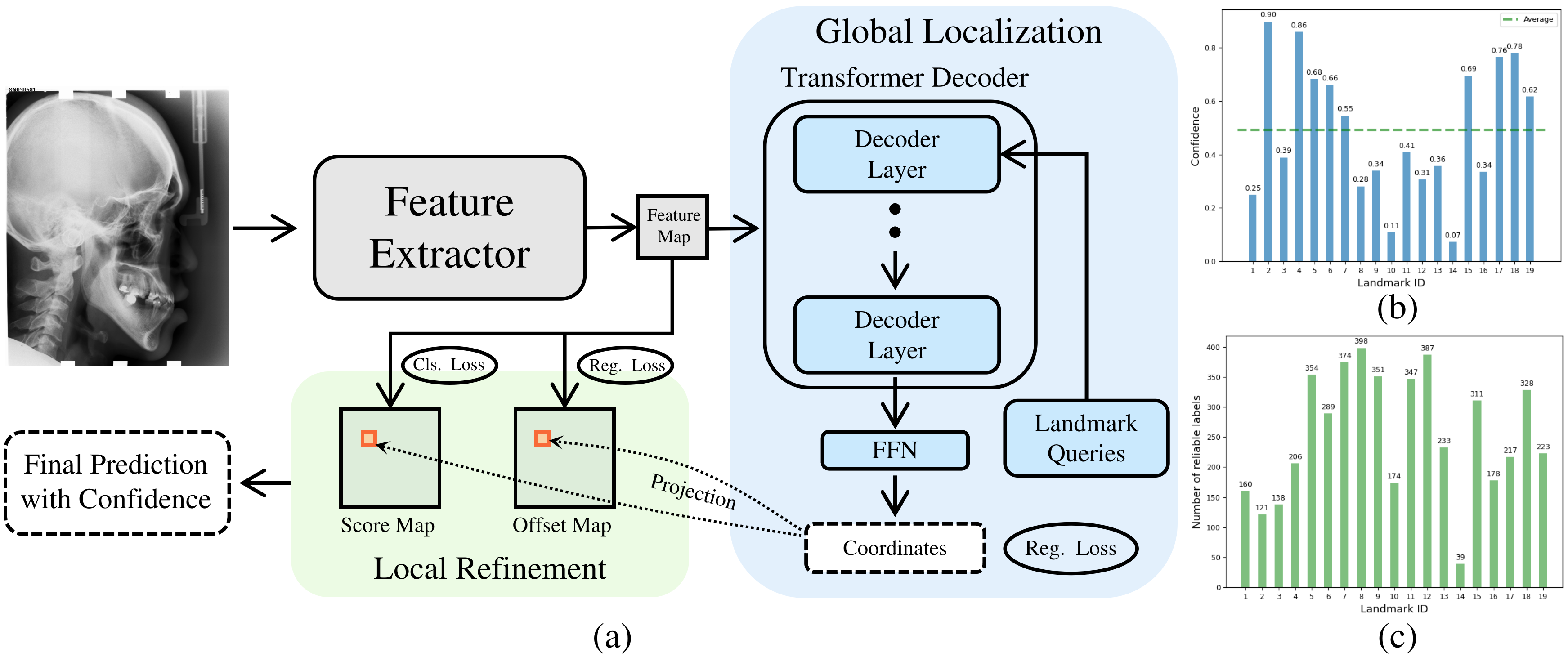}
\vspace{-2pt}
\caption{\small{(a) Our landmark detection model. (b) Confidence scores of different landmarks for a random target domain image. (c) Statistics of reliable landmark-level pseudo-labels with a fixed threshold $\tau=0.4$ over 500 images.}} 
\label{fig:base_model}    
\vspace{-2pt}  
\end{figure}

Recently, coordinate regression~\cite{LLZ20,JLW22} has obtained better performance than heatmap regression~\cite{ZLZ19,CMC19}. However, coordinate based methods do not output confidence scores, which are necessary for pseudo-label selection in self-training \cite{Lee13,BTQ21}. To address this issue, we designed a model that is able to predict accurate landmarks while providing confidence scores. As shown in Fig.~\ref{fig:base_model} (a), the model utilizes both coordinate and heatmap regression, where the former provides coarse but robust predictions via global localization, then projected to the local maps of the latter for prediction refinement and confidence measurement.   

\noindent
\textbf{Global localization.}
We adopt Transformer decoder~\cite{JLL21,LJL22} for coarse localization due to its superiority in global attentions. A convolutional neural network (CNN) is used to extract feature $f \in \mathbb{R}^{C \times H \times W}$, where $C$, $H$, and $W$ represents number of channels, map height and width, respectively. By using $f$ as memory, the decoder takes landmark queries $q \in \mathbb{R}^{L \times C}$ as input, then iteratively updates them through multiple decoder layers, where $L$ is the number of landmarks. Finally, a feed-forward network (FFN) converts the updated landmark queries to coordinates $\hat{y}_c \in \mathbb{R}^{L \times 2}$. The loss function $L_{\text{coord}}$ is defined to be the L1 loss between the predicted coordinate $\hat{y}_c$ and the label $y_c$.

\noindent
\textbf{Local refinement.}
This module outputs a score map $\hat{y}_s \in \mathbb{R}^{L \times H \times W}$ and an offset map $\hat{y}_o \in \mathbb{R}^{2L \times H \times W}$ via 1x1 convolutional layers by taking $f$ as input. The score map indicates the likelihood of each grid to be the target landmark, while the offset map represents the relative offsets of the neighbouring grids to the target. The ground-truth (GT) landmark of the score map is smoothed by a Gaussian kernel~\cite{XWW18}, and L2 loss is used for loss function $L_{\text{score}}$. Since the offset is a regression problem, L1 is used for loss $L_{\text{offset}}$, and only applied to the area where its GT score is larger than zero. During inference, different from \cite{XWW18,CMC19,LWJ20,JLS21}, the optimal local grid is not selected by the maximum score of $\hat{y}_s$, but instead the projection of the coordinates from global localization. Then the corresponding offset value is added to the optimal grid for refinement. Also, the confidence of each prediction can be easily obtained from the score map via projection.

The loss function of the landmark detection model can be summarized as    
\begin{equation}
L_{\text{base}} = \sum_{\mathcal{S}} \lambda_{\text{s}} L_{\text{score}} + \lambda_{\text{o}} L_{\text{offset}} + L_{\text{coord}}, 
\end{equation}
where $\mathcal{S}$ is source domain data, $\lambda_{\text{s}}$ and $\lambda_{\text{o}}$ are balancing coefficients. Empirically, we set $\lambda_{\text{s}}=100$ and $\lambda_{\text{o}}=0.02$ in this paper.

\subsection{Landmark-aware Self-training}
\label{sec2.2}

Self-training~\cite{Lee13} is an effective semi-supervised learning (SSL) method, which iteratively estimates and selects reliable pseudo-labeled samples to expand the labeled set. Its effectiveness has also been verified on several vision tasks under the UDA setting, such as object detection~\cite{CLS18}. However, very few works explored self-training for the UDA of landmark detection, but mostly restricted to the paradigm of SSL~\cite{DoY19,WJG22}. 

Since UDA is more challenging than SSL due to domain shift, reliable pseudo-labels should be carefully selected to avoid confirmation bias~\cite{AOA20}. Existing works \cite{DoY19,WJG22,MQH20} follow the pipeline of image classification by evaluating reliability at the image-level, which we believe is not representative because the landmarks within an image may have different reliabilities (see Fig.~\ref{fig:base_model} (b)). To avoid potential noisy labels caused by the image-level selection, we propose LAST, which selectes reliable pseudo-labels at the landmark-level. To achieve this, we use a binary mask $m \in \{0,1\}^L$ to indicate the reliability of each landmark for a given image, where value 1 indicates the label is reliable and 0 the opposite. To decide the reliability of each landmark, a common practice is to use a threshold $\tau$, where the $l$-th landmark is reliable if its confidence score $s^l > \tau$. During loss calculation, each loss term is multiplied by $m$ to mask out the unreliable landmark-level labels. Thus, the loss for LAST is
\begin{equation}
L_{\text{LAST}} = \sum_{\mathcal{S} \cup \mathcal{T}^{\prime}} M( L_{\text{base}}), 
\end{equation}
where $M$ represents the mask operation, $\mathcal{T}^{\prime}=\{x^\mathcal{T}_j,{y^\prime}^\mathcal{T}_j\}^M_{j=1}$, and ${y^\prime}^\mathcal{T}$ is the estimated pseudo-labels from the last self-training round. Note that the masks of the source domain data $\mathcal{S}$ always equal to one as they are ground truths.

However, the landmark-level selection leads to unbalanced pseudo-labels between landmarks, as shown in Fig.~\ref{fig:base_model} (c). This is caused by the fixed threshold $\tau$ in self-training, which cannot handle different landmarks adaptively. To address this issue, we introduce percentile scores~\cite{BTQ21} to yield dynamic thresholds (DT) for different landmarks. Specifically, for the $l$-th landmark, when the pseudo-labels are sorted based on confidence (high to low), $\tau^l_r$ is used as the threshold, which is the confidence score of $r$-th percentile. In this way, the selection ratio of pseudo-labels can be controlled by $r$, and the unbalanced issue can be addressed by using the same $r$ for all the landmarks. We set the curriculum to be $r=\Delta \cdot t$, where $t$ is the $t$-th self-training round and $\Delta$ is a hyperparameter that controls the pace. We use $\Delta=20\%$, which yields five training rounds in total. 

\subsection{Domain Adversarial Learning}
\label{sec2.3}
Although self-training has been shown effective, it inevitably contains bias towards source domain because its initial model is trained with source domain data only. In other words, the data distribution of target domain is different from the source domain, which is known as covariate shift~\cite{ZYZ22}. To mitigate it, we introduce DAL to align the distribution of the two by conducting an adversarial training between a domain classifier and the feature extractor. Specifically, the feature $f$ further goes through a global average pooling (GAP) and a fully connected (FC) layer, then connects to a domain classifier $D$ to discriminate the source of input $x$. The classifier can be trained with binary cross-entropy loss:
\begin{equation}
L_{\text{D}} = -d \log D(F(x)) - (1-d) \log (1-D(F(x))),
\end{equation}
where $d$ is domain label, with $d=0$ and $d=1$ indicating the images are from source and target domain, respectively. The domain classifier is trained to minimize $L_{\text{D}}$, while the feature extractor $F$ is encouraged to maximize it such that the learned feature is indistinguishable to the domain classifier. Thus, the adversarial objective function can be written as $L_{\text{DAL}}=\max\limits_{F} \min\limits_{D} L_{\text{D}}$. To simplify the optimization, we adopt gradient reversal layer (GRL)~\cite{GaL15} to mimic the adversarial training, which is placed right after the feature extractor. During backpropagation, GRL negates the gradients that pass back to the feature extractor $F$ so that $F$ is actually maximized. In this way, the adversarial training can be done via the minimization of $L_{\text{D}}$, i.e., $L_{\text{DAL}}=L_{\text{D}}$.     

Finally, we have the overall loss function as follows: 
\begin{equation}
L = \sum_{\mathcal{S} \cup \mathcal{T}^{\prime}} L_{\text{LAST}} + \lambda_{D} L_{\text{DAL}},
\end{equation}
where $\lambda_{D}$ is a balancing coefficient. 

\section{Experiments}

\subsection{Experimental Settings}

In this section, we present experiments on cephalometric landmark detection. See lung landmark detection in Appendix A.

\noindent
\textbf{Source domain.}
The ISBI 2015 Challenge provides a public dataset~\cite{WHL16}, which is widely used as a benchmark of cephalometric landmark detection. It contains 400 images in total, where 150 images are for training, 150 images are Test 1 data, and the remaining are Test 2. Each image is annotated with 19 landmarks by two experienced doctors, and the mean values of the two are used as GT. In this paper, we only use the training set as the labeled source domain data.

\noindent
\textbf{Target domain.}
The ISBI 2023 Challenge provides a new dataset~\cite{KZB23}, which was collected from 7 different imaging devices. By now, only the training set is released, which contains 700 images. For UDA setting, we randomly selected 500 images as unlabeled target domain data, and the remaining 200 images are for evaluation. The dataset provides 29 landmarks, but we only use 19 of them, i.e., the same landmarks as the source domain~\cite{WHL16}. Following previous works~\cite{LLZ20,JLW22}, all the images are resized to $640\times800$. For evaluation metric, we adopt MRE and SDR within four radius (2mm, 2.5mm, 3mm, and 4mm).

\begin{table}[t]
\centering
\footnotesize
\newcolumntype{C}{>{\centering\arraybackslash}X}%
\caption{Results on the target domain test set, under UDA setting.}
\begin{tabularx}{\textwidth}{lCCCCC}
\hline
Method & MRE$\downarrow$ & 2mm & 2.5mm & 3mm & 4mm \\ 
\hline
Base, Labeled Source & 3.32 & 50.05 & 56.87 & 62.63 & 70.87 \\
\hline
FDA~\cite{YaS20} & 2.16 & 61.28 & 69.73 & 76.34 & 84.57 \\
UMT~\cite{DLC21} & 1.98 & 63.94 & 72.52 & 78.89 & 87.05 \\
SAC~\cite{ArR21} & 1.94 & 65.68 & 73.76 & 79.63 & 87.81 \\
AT~\cite{LDM22} & 1.87 & 66.82 & 74.81 & 80.73 & 88.47 \\
\textbf{Ours} & \textbf{1.75} & \textbf{69.15} & \textbf{76.94} & \textbf{82.92} & \textbf{90.05} \\
\hline
Base, Labeled Target & 1.22 & 83.76 & 89.71 & 92.79 & 96.08 \\
\hline
\end{tabularx}
\label{tab:results_uda} 
\end{table}

\noindent
\textbf{Implementation details.}
We use ImageNet pretrained ResNet-50 as the backbone, followed by three deconvolutional layers for upsampling to stride 4~\cite{XWW18}. For Transformer decoder, three decoder layers are used, and the embedding length $C=256$. Our model has 41M parameters and 139 GFLOPs when input size is $640\times800$. The source domain images are oversampled to the same number of target domain so that the domain classifier is unbiased. Adam is used as the optimizer, and the model is trained for 720 epochs in each self-training round. The initial learning rate is $2\times10^{-4}$, and decayed by 10 at the 480th and 640th epoch. The batch size is set to 10 and $\lambda_{D}$ is set to 0.01. For data augmentation, we use random scaling, translation, rotation, occlusion, and blurring. The code was implemented with PyTorch 1.13 and trained with one RTX 3090 GPU. The training took about 54 hours.

\subsection{Results} 
For the comparison under UDA setting, several state-of-the-art UDA methods were implemented, including FDA~\cite{YaS20}, UMT~\cite{DLC21}, SAC~\cite{ArR21}, and AT~\cite{LDM22}. Additionally, the base model trained with source domain data only is included as the lower bound, and the model trained with equal amount of labeled target domain data is used as the upper bound. Tab.~\ref{tab:results_uda} shows the results. Firstly, we can see that the model trained on the target domain obtains much better performance than the one on source domain in both MRE (1.22mm vs. 3.32mm) and SDR (83.76\% vs. 50.05\%, within 2mm), which indicates that the domain shift can cause severe performance degradation. By leveraging both labeled source domain and unlabeled target domain data, our model achieves 1.75mm in MRE and 69.15\% in SDR within 2mm. It not only reduces the domain gap by a large margin (3.32mm$\rightarrow$1.75mm in MRE and 50.05\%$\rightarrow$69.15\% in 2mm SDR), but also outperforms the other UDA methods consistently. However, there is still a gap between the UDA methods and the supervised model in target domain.

\begin{figure}[t]
\centering
  \includegraphics[width=1\linewidth]{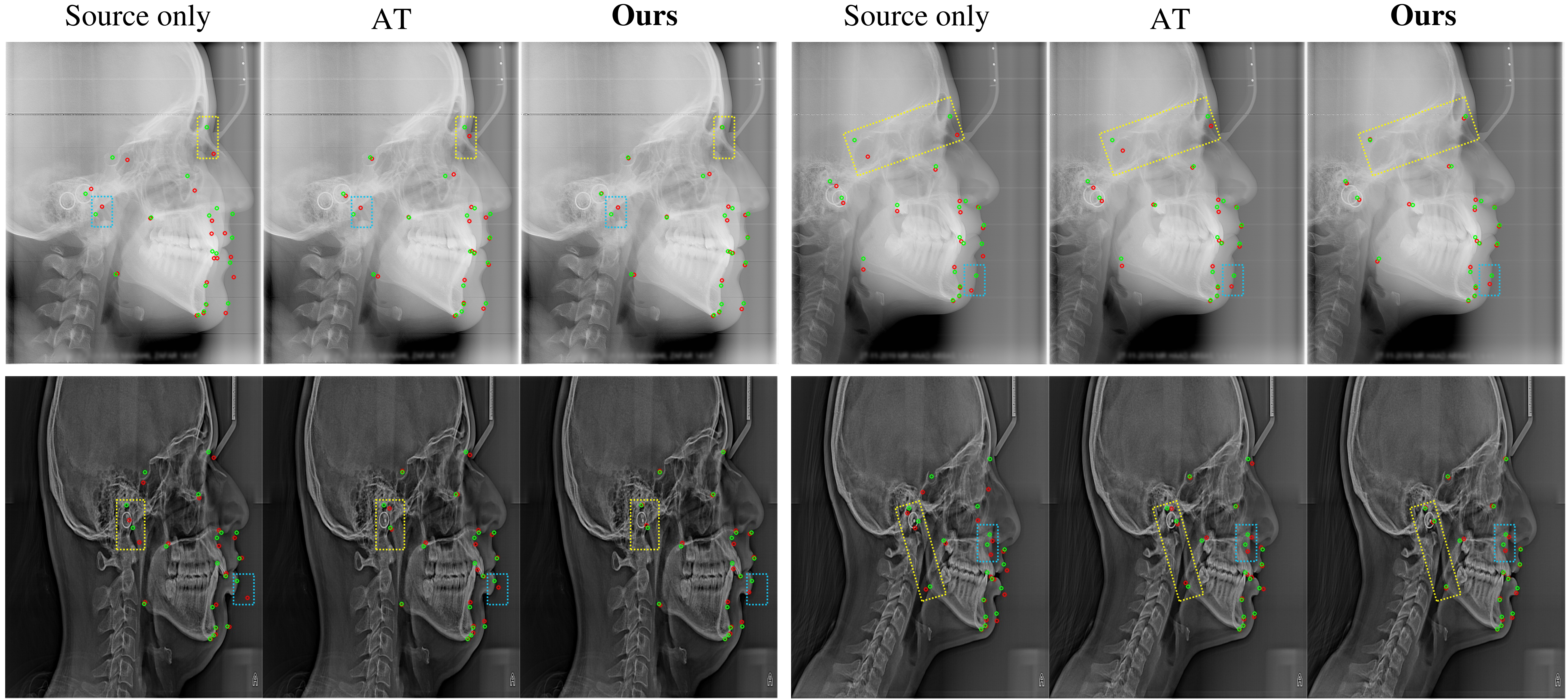}
\vspace{-2pt}
\caption{\small{Qualitative results of three models on target domain test data. \textcolor{green}{Green} dots are GTs, and \textcolor{red}{red} dots are predictions. \textcolor{yellow}{Yellow} rectangles indicate that our model performs better than the other two, while \textcolor{cyan}{cyan} rectangles indicate that all the three fail.}} 
\label{fig:vis}      
\end{figure}

\begin{table}[t]
\centering
\footnotesize
\newcolumntype{C}{>{\centering\arraybackslash}X}%
\caption{Ablation study of different modules.}
\begin{tabularx}{\textwidth}{lCCCCC}
\hline
Method & MRE$\downarrow$ & 2mm & 2.5mm & 3mm & 4mm \\ 
\hline
Self-training~\cite{Lee13} & 2.18 & 62.18 & 69.44 & 75.47 & 84.36 \\
LAST w/o DT & 1.98 & 65.34 & 72.53 & 78.03 & 86.11 \\
LAST & 1.91 & 66.21 & 74.39 & 80.23 & 88.42 \\
DAL & 1.96 & 65.92 & 74.18 & 79.73 & 87.60 \\
LAST+DAL & \textbf{1.75} & \textbf{69.15} & \textbf{76.94} & \textbf{82.92} & \textbf{90.05} \\
LAST+DAL w/ HM & 1.84 & 66.45 & 75.09 & 81.82 & 89.55 \\
\hline
\end{tabularx}
\label{tab:result_ablation} 
\end{table}

\subsection{Model Analysis}
We first do ablation study to show the effectiveness of each module, which can be seen in Tab.~\ref{tab:result_ablation}. The baseline model simply uses vanilla self-training~\cite{Lee13} for domain adaptation, which achieves 2.18mm in MRE. By adding LAST but without dynamic thresholds (DT), the MRE improves to 1.98mm. When the proposed LAST and DAL are applied separately, the MREs are 1.91mm and 1.96mm, respectively, which verifies the effectiveness of the two modules. By combining the two, the model obtains the best results in both MRE and SDR. To show the superiority of our base model, we replace it by standard heatmap regression~\cite{XWW18} (HM), which obtains degraded results in both MRE and SDR. Furthermore, we conduct analysis on subdomain discrepancy, which shows the effectiveness of our method on each subdomain (see Appendix B).

\subsection{Qualitative Results}
Fig.~\ref{fig:vis} shows the qualitative results of the source-only base model, AT~\cite{LDM22}, and our method on target domain test data. The green dots are GTs, and red dots are predictions. It can be seen from the figure that our model makes better predictions than the other two (see yellow rectangles). We also notice that some landmarks are quite challenging, where all the three fail to give accurate predictions (see cyan rectangles).

\section{Conclusion}
In this paper, we investigated anatomical landmark detection under the UDA setting. To mitigate the performance drop caused by domain shift, we proposed a unified UDA framework, which consists of a landmark detection model, a self-training strategy, and a DAL module. Based on the predictions and confidence scores from the landmark model, a self-training strategy is proposed for domain adaptation via landmark-level pseudo-labels with dynamic thresholds. Meanwhile, the model is encouraged to learn domain-invariant features via adversarial training so that the unaligned data distribution can be addressed. We constructed a UDA setting based on two anatomical datasets, where the experiments showed that our method not only reduces the domain gap by a large margin, but also outperforms other UDA methods consistently. However, a performance gap still exists between the current UDA methods and the supervised model in target domain, indicating more effective UDA methods are needed to close the gap.  

\noindent
\textbf{Acknowledgments.}
This work was supported by the Shenzhen Science and Technology Innovation Committee Fund (Project No. SGDX20210823103201011) and Hong Kong Innovation and Technology Fund (Project No. ITS/028/21FP).

%
%
%
\bibliographystyle{splncs04}
\bibliography{arxiv}

\clearpage
\appendix

\section{Experiments on Lung Landmark Detection}

We further verify the proposed method on lung landmark detection. 

\noindent
\textbf{Source domain.}
The Japanese Society of Radiological Technology (JSRT) dataset~\cite{SKI00} provides 247 CXR images with landmark annotations on lung and heart. The original resolution is $1024\times1024$, with a pixel spacing of 0.35mm. In this work, only the 94 landmarks of left and right lungs are used and the images are resized to $512\times512$.  

\noindent
\textbf{Target domain.}
\cite{GMM22} provides 94 lung landmark annotations for three CXR datasets, namely the Montgomery County dataset~\cite{CJP13} (138 images), the Shenzhen dataset~\cite{JKC13} (390 images), and the Padchest dataset~\cite{BPS20} (137 images). The above three sets are used as the target domain, where 70\% are randomly selected as unlabeled data and 30\% are test set. All the images are resized to $512\times512$. For evaluation metric, we adopt MRE and SDR within four radius (2mm, 2.5mm, 3mm, and 4mm). 

\noindent
\textbf{Implementation details.}
The implementation details of lung landmark detection mostly remain the same as that of cephalometric landmark detection, except that $\lambda_D$ is set to 0.005.

\noindent
\textbf{Results.}
Similar to the cephalometric results, we compare with state-of-the-art UDA methods, including FDA~\cite{YaS20}, UMT~\cite{DLC21}, SAC~\cite{ArR21}, and AT~\cite{LDM22}. Additionally, the base model trained with source domain data only is included as the lower bound, and the model trained with equal amount of labeled target domain data is used as the upper bound. From Tab.~\ref{tab:results_uda}, we can see that the model trained on the target domain again obtains much better performance than that of source domain. For example, the MRE of target domain is 4.52mm while the one trained on source domain is 7.55mm; the SDR within 2mm on target domain is 26.47\%  while the result trained on source domain is only 14.22\%. Therefore, the domain gap also exists in lung landmark detection. With the proposed method, we largely reduce the domain gap by further utilizing the unlabeled target domain data. Specifically, our method obtains 5.34mm in MRE and 18.10\% SDR within 2mm, which are 29\% and 27\% better than that of lower bound, respectively. Compared to other UDA methods, our method is superior in both MRE and SDRs.

\section{Subdomain Discrepancy}

\begin{table}[t]
\centering
\footnotesize
\newcolumntype{C}{>{\centering\arraybackslash}X}%
\caption{Results on the target domain test set of lung landmark detection.}
\begin{tabularx}{\textwidth}{lCCCCC}
\hline
Method & MRE$\downarrow$ & 2mm & 2.5mm & 3mm & 4mm \\ 
\hline
Base, Labeled Source & 7.55 & 14.22 & 20.55 & 26.73 & 38.73 \\
\hline
FDA~\cite{YaS20} & 6.14 & 15.51 & 22.01 & 28.82 & 42.07 \\
UMT~\cite{DLC21} & 5.82 & 16.11 & 22.92 & 30.15 & 43.87 \\
SAC~\cite{ArR21} & 5.66 & 16.69 & 23.59 & 30.86 & 45.42 \\
AT~\cite{LDM22} & 5.49 & 17.28 & 24.36 & 32.19 & 46.76 \\
\textbf{Ours} & \textbf{5.34} & \textbf{18.10} & \textbf{25.82} & \textbf{33.27} & \textbf{48.02} \\
\hline
Base, Labeled Target & 4.52 & 26.47 & 35.85 & 44.99 & 59.96 \\
\hline
\end{tabularx}
\label{tab:results_uda}
\end{table}

\begin{figure}[t]
\centering
  \includegraphics[width=1\linewidth]{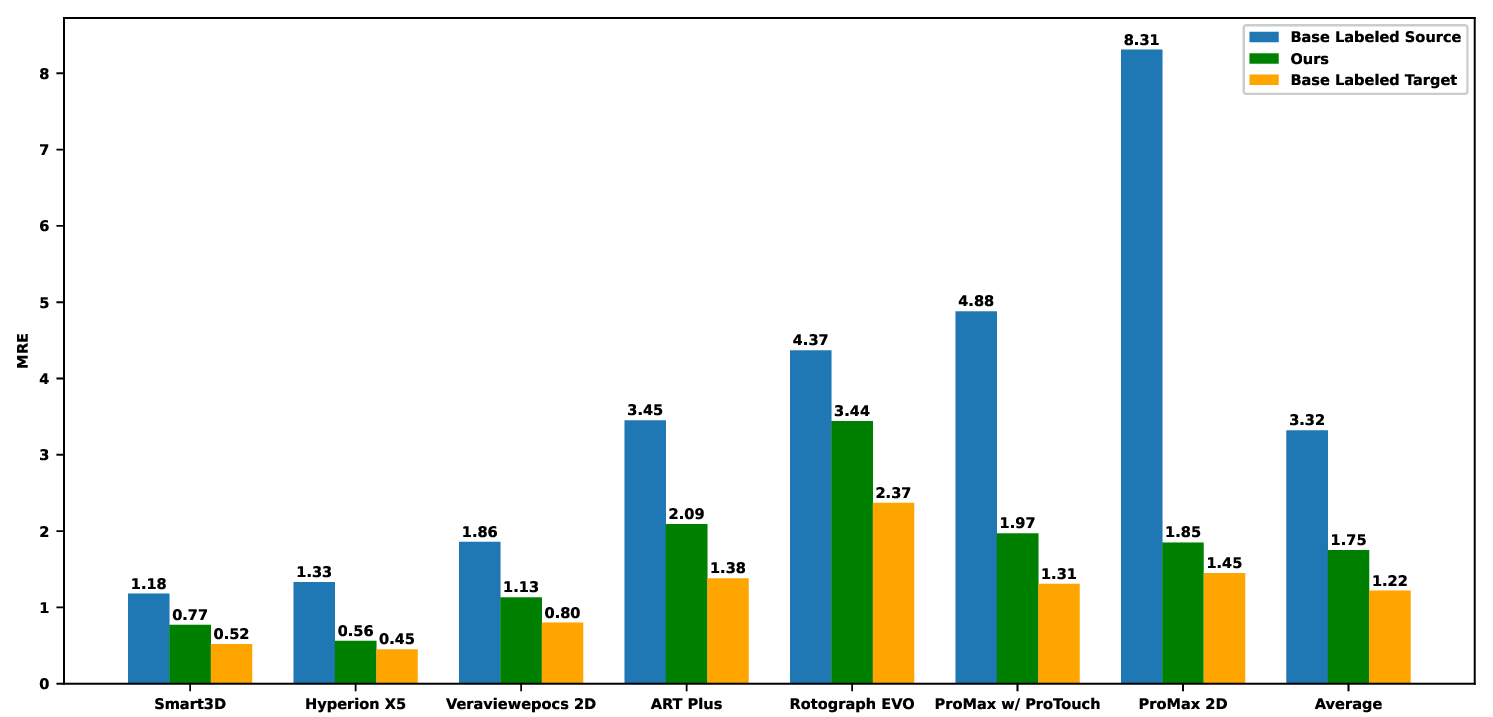}
\vspace{-2pt}
\caption{Seven subdomain results of target domain test set.} 
\label{fig:subdomain}      
\end{figure}

Since the target domain consists of images from seven different imaging devices, it would be interesting to investigate the subdomain discrepancy. Fig.~\ref{fig:subdomain} shows the subdomain results of the proposed method as well as the upper and lower bound in MRE. As can be seen, the models trained on source-only and target-only data show different characteristics in subdomain discrepancy. Specifically, the model trained on source-only data has larger subdomain discrepancy, where the ProMax 2D subset has a much larger MRE than the average (8.31mm vs. 3.32mm). In contrast, the target-only model has a relatively small subdomain discrepancy. In particular, the ProMax 2D subset of the target-only model has a small gap against the average (1.45mm vs. 1.22mm), indicating that training with similar domain images can reduce the domain gap significantly. For the proposed method, it shows similar characteristic as the target-only model, which implies the effectiveness of the proposed UDA method in reducing domain gaps by leveraging unlabeled target domain data. 

\end{document}